\documentclass[runningheads]{llncs}
\pdfoutput=1
\usepackage{graphicx}

\usepackage{tikz}
\usepackage{comment}
\usepackage{amsmath,amssymb} 
\usepackage{color}
\usepackage{bm}
\usepackage{xcolor}

\usepackage[accsupp]{axessibility}  

\begin{document}
\title{HaloAE: An HaloNet based Local Transformer Auto-Encoder for Anomaly Detection and Localization}

\titlerunning{HaloAE}

\author{Emilie Mathian\inst{1,3} \and
Huidong Liu \inst{2}\and
Lynnette Fernandez-Cuesta \inst{1} \and
Dimitris Samaras  \inst{2} \and
Matthieu Foll \inst{1} \and
Liming Chen \inst{3}
}
\authorrunning{E. Mathian et al.}

\institute{International Agency for Research on Cancer (IARC-WHO), Lyon, France \\ \email{mathiane@iarc.who.int}  \and
 Stony Brook University, New York, USA \and
 Ecole Centrale de Lyon, Ecully, France \\  \email{liming.chen@ec-lyon.fr}\\
}

\maketitle

\begin{abstract}

Unsupervised anomaly detection and localization is a crucial task as it is impossible to collect and label all possible anomalies. Many studies have emphasized the importance of integrating local and global information to achieve accurate segmentation of anomalies. To this end, there has been a growing interest in Transformer, which allows modeling long-range content interactions. However, global interactions through self attention are generally too expensive for most image scales.  In this study, we introduce HaloAE, the first auto-encoder based on a local 2D version of Transformer with HaloNet. With HaloAE, we have created a hybrid model that combines convolution and local 2D block-wise self-attention layers and jointly performs anomaly detection and segmentation through a single model. We achieved competitive results on the MVTec dataset, suggesting that vision models incorporating Transformer could benefit from a local computation of the self-attention operation, and pave the way for other applications\footnote{The code is available at: \url{https://anonymous.4open.science/r/HaloAE-8313/README.md} }.

\end{abstract}

\section{Introduction}
\begin{figure}[ht]
 \centering
 \includegraphics[width=.5\textwidth]{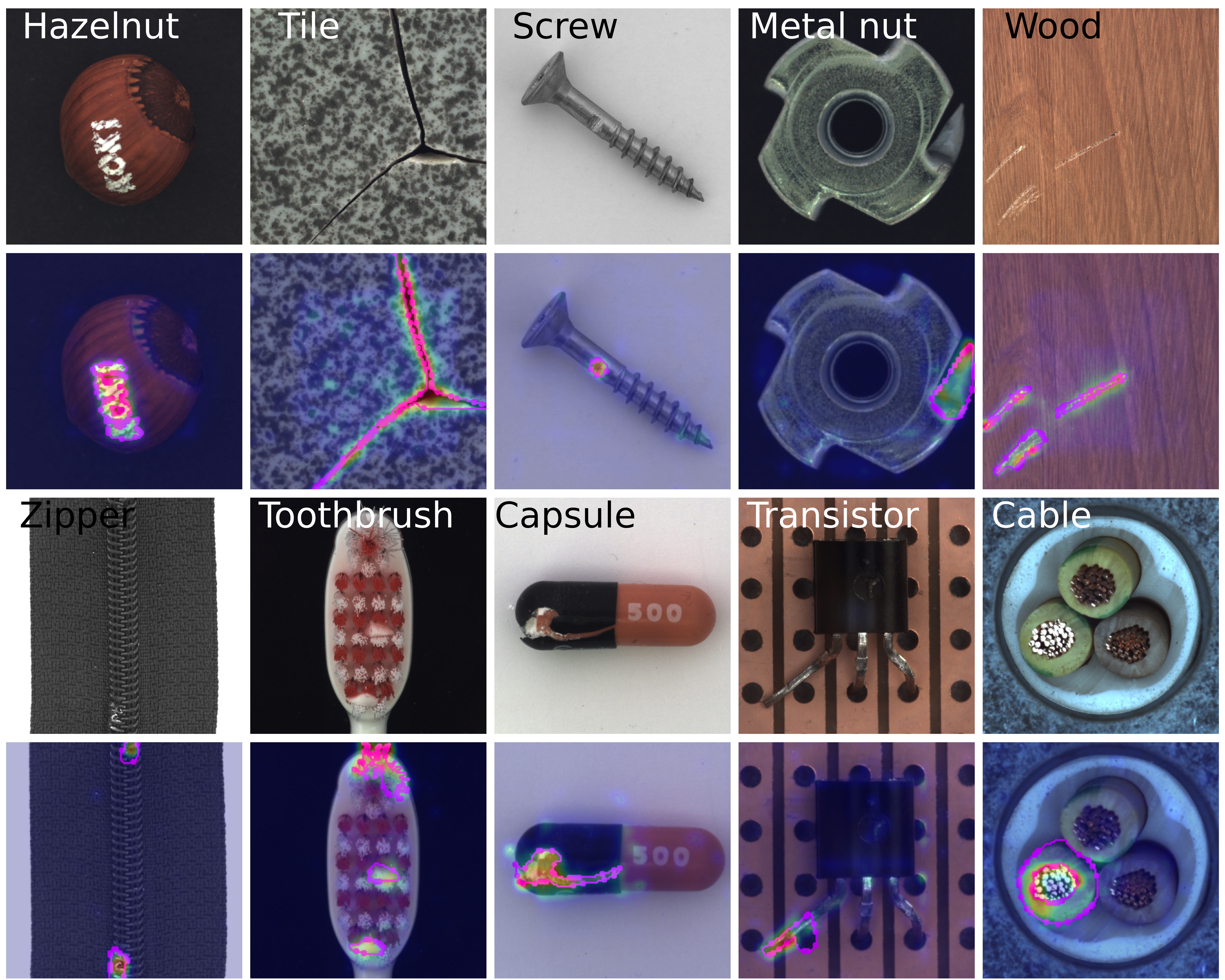}
 \caption{ Anomaly localization results from the MVTec AD dataset.  The first and third rows show the input images, the second and fourth rows show the anomaly maps generated by HaloAE, the ground truth localization is circled with a pink line.   }
 \label{fig:qualitative_res}
\end{figure}

Anomaly detection (AD) aims to determine whether a given image contains an abnormal pattern,  given a set of normal or abnormal images, while its localization or segmentation need further  to determine the subregions containing the anomalies (see Fig.\ref{fig:qualitative_res}). This is a common problem in various domains, \textit{e.g.}, in industry to detect defective objects \cite{bergmannAE2019mvtec}, \cite{bergmann2020ustudents}, \cite{dinh2016density},

in medicine \cite{schlegl2017unsupervised}, \cite{schlegl2017AnoGan}, \cite{seebock2016identifying}, or for video surveillance \cite{adam2008robust}, \cite{liu2018classifier}, \cite{sultani2018real}. Listing all anomalies is a difficult task because of their low probability density. Therefore, the problem is usually addressed via unsupervised learning approaches, also called one-class classification or out-of-distribution detection. The models use only the defect-free samples during the learning phase and attempt to identify and localize anomalies at the time of inference.

\par State of the art has featured two main approaches on AD: distribution-based or reconstruction. Distribution-based approaches generally make use of Deep Convolutional Neural Networks (Deep CNN) to extract representations of normal images or image patches and learn a parametric distribution, \textit{e.g.}, Gaussian distribution, of these deep features. They typically require to learn two models, one for anomaly detection and another for anomaly segmentation.      Reconstruction-based approaches involve training a convolutional auto-encoder  \cite{bergmannAE2019mvtec}, \cite{bergmannSSIM2018improving}, \cite{zavrtanik2021reconstructionInpainting}, \cite{zavrtanik2021draem}, or Generative Adversarial Networks (GAN) \cite{baur2018VAEGAN}, \cite{schlegl2017AnoGan}, \cite{schlegl2019fAnoGan}, \cite{akcay2018ganomaly}, \cite{akccay2019skipGanAnomaly}, \cite{venkataramanan2020attentionGAN}  to reconstruct the normal images assuming that the model should fail to reconstruct abnormal images. The advantage of such approaches is that a single model can be used for both anomaly detection and segmentation. However, most of these models \cite{bergmannAE2019mvtec}, \cite{bergmannSSIM2018improving}, \cite{schlegl2017AnoGan}, \cite{akcay2018ganomaly} do not perform well as they generalize strongly and can reconstruct anomalies.
\par Given the fact that detection of abnormal patterns requires combining local and global information, different models have been proposed, either using a fully convolutional neural network (FCNN), \cite{zhang2020multi}, or by integrating Transformer's self-attention, which models content long-range interactions \cite{vaswani2017attention}. To adapt Transformer to images, Visual Tranformer (ViT) \cite{dosovitskiy2020ViT} is typically used in an AE \cite{mishra2021vtadl}, \cite{pirnay2021inpaintTr}, \cite{yang2021SAAE}.  While CNN enables to capture easily translation invariant local patterns, capturing long-range dependencies is challenging because of CNN's poor scaling properties with respect to large receptive fields. On the other hand, Transformer enables to model dependencies between distant elements of a sequence through self-attention but the complexity of memory and computation grows quadratically with image size. Furthermore, ViT only encodes inter-patch correlations while ignoring intra-patch correlations \cite{sheynin2021localglobalTr}. 

\par In this work, we introduce HaloAE, which extends the block-based local self-attention  proposed by HaloNet  \cite{vaswani2021halonet} to AE. Our model allows us to compute intra-patch correlations unlike models using ViT \cite{dosovitskiy2020ViT}. To regularize the proposed AE and capture a multiscale and semantic rich  description of an image, multi-scale feature maps are first generated via a pre-trained network as in \cite{shi2021DFR}, \cite{yang2021SAAE},  \cite{mishra2021vtadl} and reconstructed. In addition, our model also incorporates a self-supervised learning (SSL) approach to further mitigate the generalization problem of AE. We reused the framework presented by Cut\&Paste \cite{li2021cutpaste} which defines a proxy classification task between normal and artificially damaged images, which mimic real industrial defects. The performance of HaloAE was evaluated on the challenging MvTec dataset \cite{bergmannAE2019mvtec}, an industrial dataset with 15 object classes (see Fig.\ref{fig:qualitative_res}).

The contributions of the paper are threefold:
\begin{itemize}
\item We propose HaloAE, a local self-attention-oriented AE based on HaloNet;
\item We apply the proposed AE to unsupervised anomaly detection and generate a single model for both anomaly detection and localization;
\item We show experimentally that the proposed local self-attention-based AE achieves competitive results on the MVTec benchmark.   
\end{itemize}

\section{Related Work}\label{section:related-work}
\subsection{Anomaly Detection and Localization Models}
\subsubsection{Reconstruction Based Methods:} \label{section-RW:Reconstruction}

They are the most commonly used methods for AD and localization \cite{bergmannAE2019mvtec}, \cite{bergmannSSIM2018improving}, \cite{zavrtanik2021reconstructionInpainting}, \cite{vasilev2020VAE}. They are usually based on Convolutional AE (CAE), trained to reconstruct defect-free images. At the inference time, the trained models are expected to fail to reconstruct  abnormal regions, as they differ from the observed training data. Segmentation maps of abnormal regions are obtained by per-pixel comparison between input and output images based on $L_2$ deviations \cite{bergmannAE2019mvtec}, \cite{zavrtanik2021reconstructionInpainting}, or SSIM values \cite{bergmannAE2019mvtec}. While simple and elegant in design, CAEs suffer from memory and generalize abnormal regions quite well \cite{zavrtanik2021reconstructionInpainting}, \cite{baur2018VAEGAN}. The latent space regularization enabled by variational auto-encoder (VAE) eases the identification of abnormal samples \cite{vasilev2020VAE}. Other approaches rely on the improvement of the quality of the reconstructed images using GANs \cite{baur2018VAEGAN}, \cite{schlegl2017AnoGan},\cite{akcay2018ganomaly}, \cite{akccay2019skipGanAnomaly}, \cite{venkataramanan2020attentionGAN} . 

 It has been shown that while GANs produce sharp images, they are unstable and tend to  tend to collapse when trained on a few samples \cite{baur2018VAEGAN} which can therefore generate many false positive alarms. In order to regularize the generalization capacity of the AE, DFR \cite{shi2021DFR} shows that the integration of local and global information is a key point to improve existing AD methods. They train an AE to reconstruct multi-scale feature maps, which are themselves generated by concatenating different layers of a pre-trained CNN. Our proposed method follows this line of architecture design.
\subsubsection{Distribution-based methods:}  \label{section-RW:EmbdSimilarity}
An important trend is to use large networks on external training datasets such as  ImageNet \cite{deng2009imagenet} to model the distribution of normal features. For example, P-SVDD \cite{cohen2020PSVDD}, PaDiM \cite{defard2021padim}, SPADE \cite{cohen2020SPADE}, PatchCore \cite{roth2021patchcore} or  U-Students \cite{bergmann2020ustudents}  assume that the normal data fits into a predefined kernel space.  They then have to define the distances between the normal data and the abnormal data, which are assumed to be located outside this space. To do this, P-SVDD looks for the smallest hyper-sphere that surrounds the normal data and uses the Euclidean distances to the center of the hypersphere to detect anomalies \cite{cohen2020PSVDD}.  While some models use clustering techniques \cite{cohen2020SPADE},  \cite{roth2021patchcore},  \cite{bergmannAE2019mvtec} to detect samples outside the normal distribution of the data, others model this distribution by Gaussian models \cite{defard2021padim}, \cite{gudovskiy2022cflow}, \cite{bergmann2020ustudents},  \cite{liznerski2020fcdd}.

These models \cite{defard2021padim}, \cite{cohen2020PSVDD}, \cite{cohen2020SPADE}, \cite{roth2021patchcore}, \cite{bergmann2020ustudents}, \cite{roth2021patchcore}  perform better than reconstruction-based methods but have two main limitations: first, they work with patches of images, which leads to high complexity at the time of inference, second, to measure distances, they usually use flattened descriptors that destroy spatial positional relations of 2D images. In addition, these techniques may require the use of different patch sizes, as using patches that are too large may lead the model to ignore small abnormal areas and vice versa for patches that are too small \cite{bergmann2020ustudents}.
\par In line with DFR \cite{shi2021DFR}, a number of AD methods, \textit{e.g.},  Intra \cite{pirnay2021inpaintTr}, SAAE \cite{yang2021SAAE} or PatchCore \cite{roth2021patchcore} or DifferNet \cite{rudolphdiffernet},  also explore the concept of multi-scale information gathering a step further, using Transformer or CNNs. Differnet has reused this idea of multi-scale feature maps, and implemented a normalization flow to detect anomalies \cite{rudolphdiffernet}.

\subsubsection{Self-supervised learning based methods:} \label{section-RW:SSL}
It is now widely accepted that data augmentation strategies help to regularize CNN. To this end, various inpainting reconstruction methods have been developed in the context of AD \cite{zavrtanik2021draem}, \cite{pirnay2021inpaintTr}, \cite{zavrtanik2021reconstructionInpainting}. For example Zavrtanik et al. showed that abnormal regions are less likely to be well reconstructed if they were not visible to the convolutional AE, and thus they treated this problem as a self-supervised reconstruction-by-inpaintings problem \cite{zavrtanik2021reconstructionInpainting}. However at the inference time these methods suffer from high complexity since an anomaly map is generated via a set of in-painted versions of an input image.
Many SSL-based methods have shown that the data augmentation strategy plays a critical role in defining an effective predictive task. For example, the application of basic geometric variations, such as rotations \cite{hendrycks2019geomSSL} or random affine  transformations \cite{bergman2020GeomSSL} performs poorly on texture images or symmetric objects \cite{golan2018GeomSSL}, \cite{bergman2020GeomSSL}, \cite{hendrycks2019geomSSL} \cite{li2021cutpaste}, that could be found found in the MVTec dataset \cite{bergmannAE2019mvtec}.
\par Based on this claim,  Cut\&Paste \cite{li2021cutpaste} developed an SSL model in which the proxy task is adapted to detect irregular patterns.  They created a data augmentation strategy in which a patch in an image is copied to another location after being randomly modified. This data-driven strategy outperforms the state of the art in terms of image-level classification. Nevertheless, at the time of inference, this method must use image patches to accurately locate anomalies. Our HaloAE also leverages this data augmentation strategy to regularize the proposed HaloNet-based AE.

\subsection{Visual Transformer} \label{section-RW:Tr}
Transformer \cite{vaswani2017attention} has revolutionized natural language processing (NLP) tasks, such as translation \cite{devlin2018bert}, by allowing to model distant dependencies between elements of an input sequence and to parallelize sequence processing. In the last two years, many efforts have been made to adapt Transformer to computer vision tasks, such as image classification \cite{dosovitskiy2020ViT}, \cite{liu2021swinT}, \cite{ramachandran2019sasa}, object detection \cite{carion2020Detr}, \cite{zhu2020ddter}, or image generation \cite{chen2020iGPT}, \cite{jiang2021transgan}, \cite{parmar2018imageTr}. The self-attention mechanism, at the core of Transformer, is a memory- and computationally-intensive procedure. Therefore, to allow this operation on matrices, the Transformer kernel has been redesigned according to two main approaches: either by the global or a local computation of the self-attention \cite{khan2021Surveytransformers}.
\subsubsection{Global approaches:}\label{section-RW:Global-Tr}
Visual Transformer (ViT) is the first adaptation of Transformer to images \cite{dosovitskiy2020ViT}. The architecture of ViT is very similar to the original  \cite{vaswani2017attention} but instead of taking a sequence of 1D words, 2D patches of an image are vectorized to feed a Transformer-like AE. However, this simple implementation requires a large dataset for training.  Different approaches such as  DeiT \cite{touvron2021Deit}, CrossViT \cite{chen2021crossvit} or Criss-cross \cite{huang2019crisscross} have improved the initial performance of ViT, either by reducing the amount of training data required \cite{touvron2021Deit}, or by taking into account more contextual information \cite{huang2019crisscross}, \cite{chen2021crossvit}.

However, all these methods require to decompose a 2D input into a sequence of vectors which implies two major limitations \cite{dosovitskiy2020ViT}, \cite{touvron2021Deit}, \cite{chen2021crossvit}: on the one hand this destroys intra-patch positional dependencies, and on the other hand, it does not allow the computation of correlations within patches.
\subsubsection{Local approaches:}\label{section-RW:Local-Tr}
Given these limitations, local versions of Transformer have been proposed. SASA proposed the first method based on a pure local self-attention model for images, and developed the computation of self-attention centered around each pixel via 2D grid extraction  \cite{ramachandran2019sasa}. Although promising, this method lags behind the state of the art \cite{vaswani2021halonet}. Recently, Swin Transformer has implemented a shifted window approach. It limits the self-attention operation to non-overlapping windows, while allowing connection between windows by merging neighboring patches into the deeper layers  \cite{liu2021swinT}. 

\par Vaswani et al. published HaloNet, where they proposed a block-based local self-attention  that achieved the best speed/accuracy tradeoff for image classification tasks \cite{vaswani2021halonet}. This result is obtained by violating the translation equivariance rule to obtain a better hardware utilization.  Assuming that neighboring pixels share most of their neighborhood, HaloNet extracts a local neighborhood for a block of pixels in a single run. This \textit{block-based} strategy allows parallelizing the self-attention operation. This technique improves both speed and memory management without affecting the performance, making the model more practical and hinting at its adaptation to larger widths \cite{vaswani2021halonet}.

\subsection{Transformer for anomaly detection}\label{section-RW:TrAE}
Some papers have used the advantages of self-attention for unsupervised AD \cite{zhang2020multi},  \cite{mishra2021vtadl}, \cite{yang2021SAAE}, \cite{pirnay2021inpaintTr}. For example, Zhang et al. integrated a multi-headed attention network between an AE that was adversely trained to reconstruct defect-free images \cite{zhang2020multi}. VT-ADL uses ViT \cite{dosovitskiy2020ViT} as an encoder and a CNN as a decoder to reconstruct anomaly-free images, while a Gaussian mixture density network is implemented to refine the localization of anomalies \cite{mishra2021vtadl}. Yang et al. implemented SAAE an AE based entirely on the self-attention mechanism, using ViT for both feature extraction, like DFR  \cite{shi2021DFR}, and ViT as the AE for reconstructing these multi-scale feature maps. Pirnay et al. proposed a similar architecture with InTra, which is a purely self-attention approach based on reconstruction by inpainting. They showed that the inpainting scheme can be used to hide anomalous regions to further restrict the model's ability to reconstruct them \cite{pirnay2021inpaintTr}. However, these techniques using ViT suffer from its inherent limitations. The AD is therefore conditioned by the size of the patches, by the fact that intra-patch positional information is destroyed via vectorization, and by the fact that intra-patch correlations are not taken into account, \textit{i.e.}, local information. In this work, we propose to leverage a block-based local attention, \textit{i.e.}, HaloNet, to define our AE to acheive a single model for both anomaly detection and segmentation \cite{vaswani2021halonet}.

\section{Method}\label{section:method}

 \begin{figure}[ht]
 \centering
 \includegraphics[width=\textwidth]{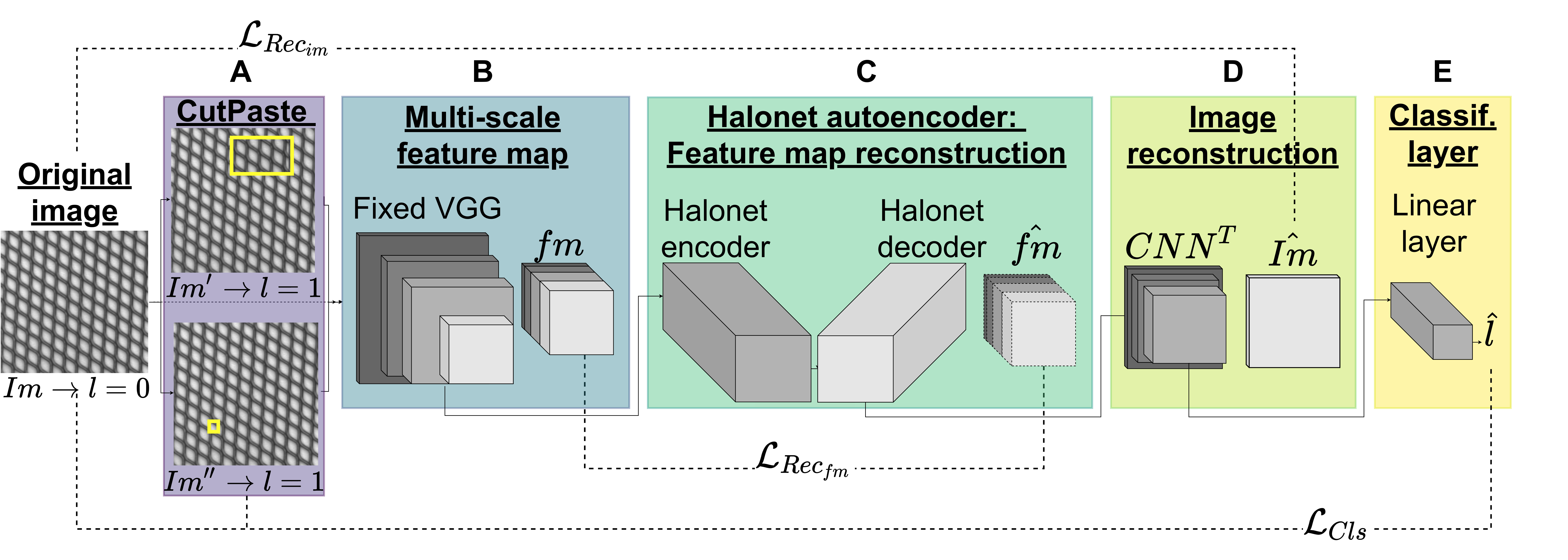}
 \caption{ Overview of HaloAE for AD. \textbf{A)} Cut\&Paste data augmentation strategy for the SSL \cite{li2021cutpaste}. \textbf{B)} Multi-scaled feature map extraction via a pretrained VGG19 network \cite{simonyan2014vgg19} on ImageNet \cite{deng2009imagenet}. \textbf{C)} Halonet AE for feature map reconstruction. \textbf{D)} Reconstruction of images via transposed VGG blocks. \textbf{E)} Linear layer to determine the classification loss. $Im$ and $\hat{Im}$, refer to the original image and the reconstructed image respectively, similarly $fm$ and $\hat{fm}$ refer to the feature map and its reconstruction. $l$ and $\hat{l}$ refer to the label and its prediction, $0$ is associated to the original picture, and $1$ to its augmented versions. $\mathcal{L}_{cls}$, $\mathcal{L}_{Rec_{fm}}$ and $\mathcal{L}_{Rec_{im}}$ refer to the classification loss and reconstruction quality of feature maps and images respectively.}
 \label{figMainArchi}
\end{figure}
\subsection{Architecture}
\subsubsection{Self-supervised learning framework:} \label{section-method:SSL}
 To mimic industrial anomalies on the MVTec dataset \cite{bergmannAE2019mvtec} we re-used the strategy set up by Cut\&Paste. This involves cutting out a rectangular patch, or any other shape according to the \textit{scar} strategy of Cut\&Paste. These elements of variable size and aspect ratio are cut from an input image and re-pasted at a random location, after undergoing random transformations such as rotations or color variations (Fig.\ref{figMainArchi} - block A). This framework allows to define a proxy 2-ways classification task between normal and artificially damaged images. Let $IM$ be the set of training images of size $N$ such that  $IM = \{im_0, ..., im_N\}$, where each $im_i$  is in $\mathbb{R} ^ {w \times h \times c}$, with $w$, $h$ and $c$  the input width, height and number of channels. We define a classification loss function such as: 

\begin{align}
    \mathcal{L}_{cls} =  \frac{1}{N} \sum_{i=0}^N \mathbb{CE}(g(\hat{im_i}), l=0) + \mathbb{CE}(g(CP(\hat{im_i})), l=1),
    \label{eq:CPLoss}
\end{align}
where the function $\mathbb{CE}$(.) refers to the binary cross entropy function, $CP(.)$ to the Cut\&Paste data augmentation strategy, and $g(.)$ to the binary classifiers, shown  in Fig.\ref{figMainArchi} - block E. This terminal linear layer takes as input a reconstructed image $\hat{im_i}$ associated with its label $l$ , which is equal to 1 if the image has been augmented by $CP(.)$ and $0$ otherwise.

\subsubsection{Image features extraction:} \label{section-method:FMGeneration}
Following the DFR \cite{shi2021DFR} method, we used a VGG19 \cite{simonyan2014vgg19} network trained on ImageNet \cite{deng2009imagenet} to extract the multi-scale features of an image. 
Given a deep CNN composed of $L$ layers,  it generates $L$ feature maps, denoted $\{\Phi_1(im), ...,\Phi_L(im)\}$, with lower layer feature maps encoding low-level patterns such as texture with small receptive fields, while deep layer feature maps capturing higher-level information such as objectness with larger receptive fields. To bring all these pieces of multi-scale information together, we aggregate these different feature maps from $\{\Phi_l(im)\}_{l=1}^L$ to achieve a multi-scale feature map. 
Since VGG19 is composed of 16 convolution layers, we choose to concatenate the $1^{st}$, $3^{rd}$, $5^{th}$, and $8^{th}$ layers of this network, which have a receptive field of 3, 10, 24, and 48 pixels, respectively. As reported by PathCore \cite{roth2021patchcore}, we exclude features from very deep layers to avoid using overly generic features that are heavily biased towards ImageNet classification. The resulting multi-scale feature map is denoted $fm \in \mathbb{R}^{w_1 \times h_1 \times c_1}$, here $w_1$ and $h_1$  equal to  $64$ and $c_1$  to 704, (see Fig.\ref{figMainArchi} - block B).

\subsubsection{Reconstruction strategy grounded on Halonet:} \label{section-method:Reconstruction}
\par The self-attention operation captures distant relationships between pixels and generates spatially varying filters unlike convolutional layers \cite{ramachandran2019sasa}, \cite{vaswani2021halonet}. We make use of the block-based local self-attention introduced by HaloNet to create a reconstruction of  $fm$ denoted $\hat{fm}$. $fm$ is divided into a grid of non overlapping blocks of size $\frac{h_1}{b},\frac{w_1}{b} $. Every block behaves like a group of query pixels. The haloing operation combines bands of $hl$ pixels around each block to obtain the shared neighborhood from which the keys and values are calculated. In this way, the local self-attention per block multiplies each pixel in a shared neighborhood, after they have been transformed by the same linear transformation, by a probability considering both content-content and content-geometry interactions, resulting in spatially varying weights  (\cite{vaswani2021halonet} eq.2 and  eq.3). In our study design we set the block size $b$ to $12$ and $hl$ to 2, instead of using the original values which are 8 and 3 respectively, in order to capture more contextual information by taking advantage of the reduced size of the input since  $h_1 = h/4$. 
\par The architecture proposed by Vaswani et al. \cite{vaswani2021halonet} is modified while  keeping its ResNets-like  structure \cite{he2016resnet}. Specifically, we have modified: (a) the head layer, substituting the 7x7 convolution with a stride of 2 by a 5x5 convolutional layer with a stride of 1, so as not to reduce the spatial dimension of the input map again; (b) the number of blocks per stage is set at 1 instead of the 3 or 4 in the original architecture, in order to create  a lighter memory model;  (c) in each block the second $1 \times 1$ convolution is replaced  by a convolution layer with a filter of size $3 \times 3$ for the first two stages and $5 \times 5$ for the last two. This last modification allows both extracting local information with the 2D convolution layer. Increasing the filter size in the last two steps avoids the blocking effect created by HaloNet that could decrease the quality of the reconstruction (Table \ref{Table:HaloArchi}). It is important to note that we don't reduce the width and height of the input feature map since this reduction has already been performed by the pre-trained VGG19 network \cite{simonyan2014vgg19}, therefore the HaloNet encoder learns a compressed version of the feature map $fm$ by reducing its channels count. The encoded feature map $fm_{enc}$ is in $\mathbb{R}^{60 \times 60 \times 58}$. From these encoded features $fm$ is reconstructed by decoder combining both convolutional and block-local self attention layers. From these encoded features, fm is reconstructed by decoder combining both convolutional layers and local block self-attention layers. The decoder
follows a similar architecture as the encoder, but all convolutional layers have been replaced by transposed convolutional layers, so we proposed the first transposed Halonet version.

\begin{table}[ht] 
\centering
\caption{Summary of the HaloNet AE architecture: Each brace encloses a block, the number of  blocks per stage is indicated in front of it. The batch normalization operation is denoted by BN, the convolution layers and the transposed convolution layers are denoted by conv and convT respectively. Finally, the number of channels at the end of each stage is indicated in the right-hand column for the encoder and decoder. }
\label{Table:HaloArchi}
\begin{tabular}{|ll|ll|}
\hline
\multicolumn{2}{|c|}{\textbf{Halonet encoder}}  & \multicolumn{2}{c|}{\textbf{Halonet decoder}}  \\ \hline
\multicolumn{1}{|c|}{$ 5 \times 5 \mbox{ conv, BN, relu} $}   & \multicolumn{1}{|c|}{$d_h = 704$}   & \multicolumn{1}{l|}{$3\times\left\{\begin{array}{l}
1 \times 1 \mbox{ conv\textsuperscript{T}, BN} \\
\mbox{Attention}(b,h), \mbox{relu}\\
3 \times 3 \mbox{ conv\textsuperscript{T}, BN}
\end{array}
\right.$} &  
$\begin{array}{l}
d_{dec_{s1}} = 29 \\
d_{dec_{s2}} =  55 \\
d_{dec_{s3}} = 118
\end{array}$
\\ \hline
\multicolumn{1}{|l|}{
$2\times\left\{
\begin{array}{l}
1 \times 1 \mbox{ conv, BN} \\
\mbox{Attention}(b,h) , \mbox{relu}\\
3 \times 3 \mbox{ conv, BN}
\end{array}
\right.$
} 
& 
$\begin{array}{l}
d_{enc_{s1}} = 234 \\
d_{enc_{s2}} = 117 \\
\end{array}$
& 
\multicolumn{1}{l|}{
$1\times
\left\{\begin{array}{l}
1 \times 1 \mbox{ conv\textsuperscript{T}, BN} \\
\mbox{Attention}(b,h), \mbox{relu}\\
5 \times 5 \mbox{ conv\textsuperscript{T}, BN}
\end{array}
\right.$
} 
&                                            
$\begin{array}{l}
d_{dec_{s4}} = 237 \\
\end{array}$
\\ \hline
\multicolumn{1}{|l|}{
$2\times\left\{
\begin{array}{l}
1 \times 1 \mbox{ conv, BN} \\
\mbox{Attention}(b,h) , \mbox{relu}\\
5 \times 5 \mbox{ conv, BN}\end{array}
\right.$
}       

& 
$\begin{array}{l}
d_{enc_{s3}} = 58 \\
d_{enc_{s4}} = 29 \\
\end{array}$
& 
\multicolumn{1}{l|}{
$1\times\left\{
\begin{array}{l}
1 \times 1 \mbox{ conv\textsuperscript{T}, BN} \\
\mbox{Attention}(b,h), \mbox{relu}\\
1 \times 1 \mbox{ conv\textsuperscript{T}, BN}
\end{array}
\right.$
}
&       
$\begin{array}{l}
d_{dec_{s5}} = 704 \\
\end{array}$
\\ \hline
\end{tabular}
\end{table}

The quality of the reconstructed feature maps is evaluated by a per-pixel loss $L_2$ and by a perceptual loss called the structure similarity index $SSIM$ \cite{bergmannSSIM2018improving}. Unlike $L_2$ which assumes independence between neighboring pixels, the $SSIM$ index evaluates the structural differences between the regions of the original and reconstructed maps by taking into account the co-variance between the regions \cite{zavrtanik2021reconstructionInpainting}, \cite{mishra2021vtadl}. Therefore, the loss associated with feature map reconstruction is given by:
\begin{align}
    \mathcal{L}_{Rec_{fm}} =  \sum_{i=1}^{h_1} \sum_{j=1}^{w_1} ||fm_{i,j} - \hat{fm}_{i,j}||_2 + (1 - SSIM(fm, \hat{fm}))_{(i,j)} ,
    \label{eq:Loss_rec_fm}
\end{align}
where the $SSIM$ is calculted between patches centered at $(i,j)$.

\par To obtain a refined anomaly map at the image scale, we implemented a small transposed convolutional neural network, which is trained to reconstruct the input image $im$ from $\hat{fm}$. It consists of five 2D convolution layers with filters of size $3\times3$, followed by a $ReLU$ activation function. Finally, a layer using the 2D nearest neighbor method is used to oversample the reconstructed image $\hat{im}$  to the scale of $im$ (Fig.\ref{figMainArchi} - block D). We use the same equation as eq.\ref{eq:Loss_rec_fm} for the reconstruction of $\hat{im}$ as follows:
\begin{align}
    \mathcal{L}_{Rec_{im}} =  \sum_{i=1}^{h} \sum_{j=1}^{w} ||im_{i,j} - \hat{im}_{i,j}||_2 + (1 - SSIM(im, \hat{im}))_{(i,j)}.
    \label{eq:Loss_rec_im}
\end{align}

\subsection{Loss function} \label{section-method:Loss}

By combining the losses described by eqs. \ref{eq:CPLoss}, \ref{eq:Loss_rec_fm}, and  \ref{eq:Loss_rec_im}, we are defining a multi-objective problem. Usually the total loss $\mathcal{L}_T$ is written as a linear combination of the different losses $\mathcal{L}_i$ such that:
\begin{align}
    \mathcal{L}_{T} =  \sum_{i} \alpha_i \mathcal{L}_i + \mathbb{R}(\bm{\alpha}),
    \label{eq:Loss_linear_comb}
\end{align}
where $\bm{\alpha}$ denotes a set of weights and $\mathbb{R}(.)$ some regularization of these weights. In general, the individual terms are weighted equally, assuming that each task contribute equally to the total loss, or $\bm{\alpha}$ weights are adjusted indivudally using an extensive grid search  \cite{groenendijk2021alphamulti}. It has been shown that the exact value chosen for these weights can strongly affect the performance of the models \cite{groenendijk2021alphamulti} \cite{dosovitskiy2019yoto}. This can be explained by the fact that some losses might be in conflict, like in our case the classification loss and the reconstruction losses, while other can benefit from each other like the $L_2$ term and the SSIM term in our reconstruction equations (see eq.\ref{eq:Loss_rec_fm} and eq.\ref{eq:Loss_rec_im}).  Inspired by the fact that humans often learn from an easy concept to a more difficult one, as pointed out by Li et al \cite{li2017self}, we implemented an adaptive weighting of the total loss function during learning.  Therefore, the weighting of different $\mathcal{L}_T$ terms changes with the number of epochs $t$ such that:

\begin{align}
    \mathcal{L}_{T}(t) =   \alpha_{1}(t) \mathcal{L}_{cls} + \alpha_{2}(t) \mathcal{L}_{Rec_{fm}} + \alpha_{3}(t) \mathcal{L}_{Rec_{im}}.
    \label{eq:Loss_total}
\end{align}
\par We assume that the classification task is easier compared to the two reconstruction tasks, since it is a global decision at the image level while the quality of the reconstructions is evaluated at the pixel level. Moreover, since the quality $\hat{im}$ depends on the quality of $\hat{fm}$, we assume that  $\mathcal{L}_{Rec_{fm}}$ must be optimized before $\mathcal{L}_{Rec_{im}}$. To this end, we modeled the evolution of  $\alpha_1$ by a decreasing logistic function, and $\alpha_2$ and $\alpha_3$ by two increasing logistic functions lagged by the number of epochs.

 \begin{figure}[ht]
 \centering
 \includegraphics[height=.3\textwidth]{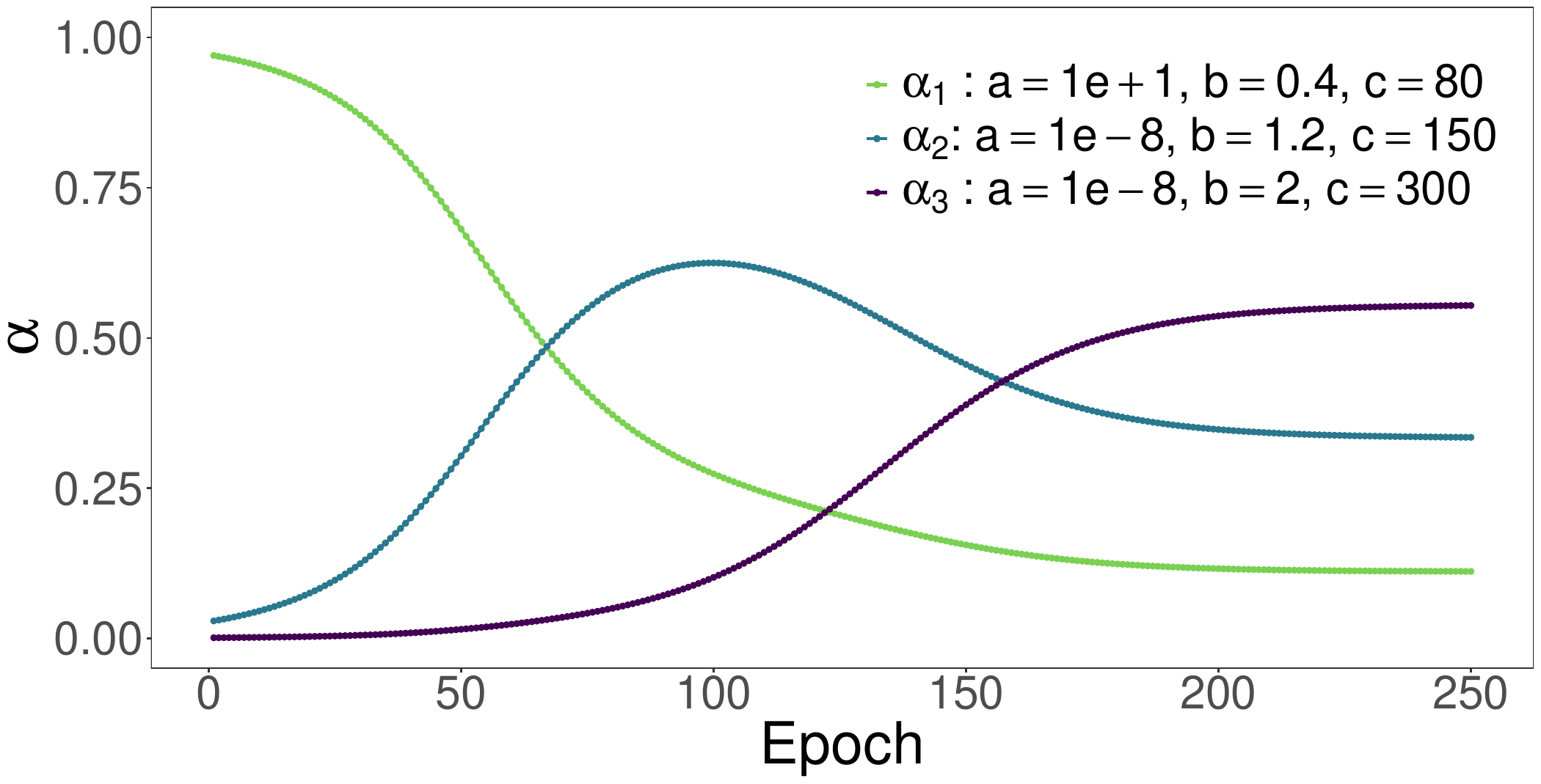}
 \caption{Evolution of $\bm{\alpha}$ along with the number of epochs. Each curve is modeled according to the following equation, whose parameters are indicated in the legend: $\frac{(a-b)}{1+\exp(x-\frac{c}{2})^{0.05}} + b$. The $\bm{\alpha}$ values are then normalized so that they sum to 1.} 
 \label{figAlpha}
\end{figure}
 \section{Experiments}\label{section:experiments}
 
 \subsection{Experimental set-up}
 We evaluated our model on the recent challenging MVTec AD dataset \cite{bergmannAE2019mvtec}. The MVTec images were resized to $256\times256$ pixels. We applied data augmentation by randomly modulating the color. As explained above, each image is associated with two artificially damaged images using the Cut\&Paste approach, either by copying and pasting a rectangle or a random shape \cite{li2021cutpaste}. 
 
 All models have the same training hyperparameters: 250 training epochs, an Adam type optimizer with a learning rate of $1e^{-4}$ and a weight decay of $1e^{-5}$,  the batch size is 12. 
 
 \par To assess our method, we calculated anomaly maps by comparing $fm$ and its reconstruction $\hat{fm}$ via the  $L_2$ distance such that:  
 \begin{align}
    A_{fm} =  \sum_{i=1}^{h_1} \sum_{j=1}^{w_1} ||fm_{i,j} - \hat{fm_{i,j} } ||_2.
    \label{eq:anom_map}
\end{align}
To obtain an anomaly map from $fm$ at the scale of $im$, we upsampled them by linear interpolation. Empirically, we observed that $A_{fm}$ gives the best results in terms of both classification and localization tasks. The classification scores according to $\mathcal{L}_{cls}$ values and the segmentation scores obtained with $A_{im}$ together with their associated anomaly maps resulting from image reconstructions, are given in supplementary Table \ref{supTableDetailScores}. 

\par The anomaly maps were post-processed to improve results (supplementary Fig. \ref{figPostProcess}). They are first normalized by the average anomaly map of the training data and denoted  $A_{fm_N}$.  All the $N$ anomaly maps from the training set $A_{fm_j}$ are then concatenated before being averaged along the channel axis of dimension $c_1$. This operation reduces potential noise (supplementary Fig.\ref{figPostProcess}). Finally, the anomaly maps are filtered using a Gaussian kernel of size $3\times3$, which smoothes the boundaries of the anomalous regions (Supplementary Fig.\ref{figPostProcess}). Image-level AD is reported by the threshold-agnostic ROC-AUC metric. For the localization we reported the pixel-wise ROC-AUC.

 \subsection{Quantitative results}
 We compared our method with alternatives, ranging from  AE $L_2$ \cite{bergmannAE2019mvtec}, which can be considered the simplest model, to DifferNet \cite{rudolphdiffernet}, which uses the notion of likelihood through normalization flow to detect anomalies. We have also included Cut\&Paste  \cite{li2021cutpaste} and DFR \cite{shi2021DFR} since we have reused parts of their method. Note that we recomputed the DFR results to have both the image-level AD ROC-AUC metric and the per-pixel segmentation ROC-AUC scores that are not available in the original paper.  We also included P-SVDD  \cite{cohen2020PSVDD} to refer to an embedding-similarity based method, as well as SAAE \cite{yang2021SAAE} and  InTra \cite{pirnay2021inpaintTr} to refer to two other techniques using Transformer. The results are summarized in Table \ref{TableMainClassifSeg}. 
 
 \par We can observe that HaloAE obtains satisfactory results for the detection of anomalies at the image level with an average ROC-AUC score of \textbf{91.4\%}. This result is strongly influenced by the poor performance obtained specifically on the carpet object. As illustrated in Fig. \ref{figResCarpet}, the network seems to be able to reconstruct the anomalies for this object, thus the distribution of  $A_{fm}$ means is similar between normal and abnormal objects. Notably, for some objects, image-level classification results are improved using the classification loss term $\mathcal{L}_{cls}$ (Table \ref{supTableDetailScores}), although on average the post-processed anomaly map computed on the feature maps gives better results. For the pixel-wise segmentation results, HaloAE obtains an average ROC-AUC score of  \textbf{91.2\%}. This score is negatively impacted by  the tile object, as shown in  Fig.\ref{figResTile}, HaloAE tends to detect the contours of anomalies. This can be explained by the local attention mechanism which calculates the interaction between neighboring pixels with respect to a central pixel.
 
 \par It is important to note that P-SVDD \cite{cohen2020PSVDD}, InTra \cite{pirnay2021inpaintTr} and Cut\&Paste \cite{li2021cutpaste} are patch-based methods, either for learning and inference \cite{cohen2020PSVDD} or for inference only \cite{li2021cutpaste}, \cite{pirnay2021inpaintTr}. Moreover, Cut\&Paste uses two different models, one for classification and one for segmentation, while we proposed an all-in-one method.  The preprocessing technique of InTra differs according to the objects, so this model is not strictly unsupervised. Finally, SAAE does not report classification scores, which are not necessarily correlated with segmentation scores, as illustrated by our results on carpet.
\begin{table}[ht]
 \caption{Anomaly detection and localization performance on the MVTec dataset. The first score in the pair refers to the image-level AD ROC-AUC score in percent, and the second to the pixel-wise ROC-AUC score in percent. The best score for each object is highlighted in bold.  }
 \label{TableMainClassifSeg} 
 \scalebox{0.72}{
\begin{tabular}{l|c|c|c|c|c|c|c|c|}
\cline{2-9}
\multicolumn{1}{c|}{\textbf{}}            & \textbf{AE-l2} & \textbf{P-SVDD}                        & \textbf{DFR}                           & \textbf{Cut\&Paste}                                              & \textbf{SAAE}                       & \textbf{InTra}                                                     & \textbf{DifferNet}                  & \textbf{HaloAE}                                                  \\ \hline
\multicolumn{1}{|l|}{Carpet}              & (-, 59.0)      & (92.9, 92.6)                           & (95.6, 98.5) & (\textbf{100.0}, 98.3)                          & (-, 97.9)                           & (98.8, \textbf{99.2})                                                       & (84.0, -)                           & (69.7, 89.4)                                                     \\
\multicolumn{1}{|l|}{Grid}                & (-, 90.0)      & (94.4, 96.2)                           & (95.0, 97.4)                           & (96.2, 97.5)                                                     & (-, 97.9)                           & (\textbf{100.0}, \textbf{98.8})  & (97.1, -)                           & (95.1, 83.1)                                                     \\
\multicolumn{1}{|l|}{Leather}             & (-, 75.0)      & (90.9, 97.4)                           & (99.4, 99.3)                           & (95.4, 99.5)                                                     & (-, 99.6)                           & (\textbf{100.0}, \textbf{99.5}) & (99.4, -)                           & (97.8, 98.5)                                                     \\
\multicolumn{1}{|l|}{Tile}                & (-, 51.0)      & (97.8, 91.4)                           & (93.1, 90.9)                           & (\textbf{100.0}, 90.5)                          & (-, \textbf{97.3}) & (98.2, 94.4)                                                       & (92.9, -)                           & (95.7, 78.5 )                                                    \\
\multicolumn{1}{|l|}{Wood}                & (-, 73.0)      & (96.5, 90.8)                           & (98.9, 95.4)                           & (99.1, 95.5)                                                     & (-, \textbf{97.6}) & (97.5, 88.7)                                                       & (99.8, -)                           & (\textbf{100.0}, 91.1)                          \\ \hline
\multicolumn{1}{|l|}{\textbf{Mean Text.}} & (-, 69.6)      & (94.5, 93.7)                           & (96.4, 96.3)                           & (98.1, 96.3)                                                     & (-, \textbf{98.2}) & (\textbf{98.9}, 96.1)                             & (94.6, -)                           & (89.7, 88.1)                                                     \\ \hline
\multicolumn{1}{|l|}{Bottle}              & (-, 86.0)      & (98.6, 98.1)                           & (99.8, 95.8)                           & (99.9, 97.6)                                                     & (-, 97.9)                           & (\textbf{100.0}, \textbf{97.1})  & (99.0, -)                           & (\textbf{100.0}, 91.9)                          \\
\multicolumn{1}{|l|}{Cable}               & (-, 86.0)      & (90.3, \textbf{96.8}) & (78.9, 91.4)                           & (\textbf{100.0}, 90.0)                          & (-, \textbf{96.8}) & (70.3, 91.0)                                                       & (86.9, -)                           & (84.6, 87.6)                                                     \\ \cline{1-1}
\multicolumn{1}{|l|}{Capsule}             & (-, 88.0)      & (76.7, 95.8)                           & (96.2, \textbf{98.5}) & (\textbf{98.6}, 97.4)                           & (-, 98.2)                           & (86.5, 97.7)                                                       & (88.8, -)                           & (88.4, 97.8)                                                     \\
\multicolumn{1}{|l|}{HazelNut}            & (-, 95.0)      & (92.0, 97.5)                           & (97.0, 92.0)                           & (93.3, 97.3)                                                     & (-, 98.5)                           & (95.7, \textbf{98.3})                                                       & (91.1, -)                           & (\textbf{99.8}, 97.8) \\
\multicolumn{1}{|l|}{MeatalNut}           & (-, 86.0)      & (94.0, \textbf{98.0}) & (93.1, 93.3)                           & (86.6, 93.1)                                                     & (-, 97.6)                           & (\textbf{96.9}, 93.3)                             & (95.1, -)                           & (88.4, 85.2)                                                     \\
\multicolumn{1}{|l|}{Pill}                & (-, 85.0)      & (86.1, 95.1)                           & (91.9, 96.8)                           & (\textbf{99.8}, 95.7)                           & (-, 98.1) & (90.2, \textbf{98.3})                                                       & (95.9, -)                           & (90.1, 91.5)                                                     \\
\multicolumn{1}{|l|}{Screw}               & (-, 96.0)      & (81.3, 95.7)                           & (94.3, \textbf{99.0}) & (90.7, 96.7)                                                     & (-, 98.9)                           & (95.7,\textbf{ 99.5})                                                      & (99.3, -) & (89.6, \textbf{99.0})                           \\
\multicolumn{1}{|l|}{Toothbrush}          & (-, 93.0)      & (100.0, 98.1)                          & (100.0, 98.5)                          & (97.5, 98.1)                                                     & (-, 98.1)                           & (\textbf{100.0}, \textbf{98.9}) & (96.1, -)                           & (97.2, 92.9)                                                     \\
\multicolumn{1}{|l|}{Transistor}          & (-, 86.0)      & (91.5, \textbf{97.0}) & (80.6, 79.1)                           & (\textbf{99.8}, 93.0)                           & (-, 96.0)                           & (95.8, 96.1)                                                       & (96.3, -)                           & (84.4, 87.5)                                                     \\
\multicolumn{1}{|l|}{Zipper}              & (-, 77.0)      & (97.9, 95.1)                           & (89.9, 96.9)                           & (\textbf{99.9}, \textbf{99.3}) & (-, 96.9)                           & (99.4, 99.2)                                                       & (98.6, -)                           & (99.7, 96.0)                                                     \\ \hline
\multicolumn{1}{|l|}{\textbf{Mean Obj.}}  & (-, 87.8)      & (90.8, 96.7)                           & (91.6, 94.4)                           & (\textbf{96.6}, 95.8)                           & (-, \textbf{97.7}) & (93.1, 96.9)                                                       & (94.6, -)                           & (92.2, 92.7)                                                     \\ \hline
\multicolumn{1}{|l|}{\textbf{Mean}}       & (71.0, 82.9)   & (92.1, 95.7)                           & (93.3, 95.1)                           & (\textbf{97.1}, 96.0)                           & (-, \textbf{97.9}) & (95.0, 96.7)                                                       & (94.9, -)                           & (91.4 91.2)                                                      \\ \hline
\end{tabular}}
\end{table}

  \subsection{Qualitative results}
  
We visualize some results of the anomaly localization in Fig. \ref{fig:qualitative_res}. The first and third rows show the input images while the second and last rows show the post-processed anomaly maps. These representations highlight that HaloAE is capable of locating tiny defects, as illustrated by the screw, capsule or zipper, and large defects, as illustrated by the hazelnut or the tile. In addition, HaloAE detects both structural defects, as shown by the wood and the tile, and color defects, as in the cable example, where the cable in the lower left corner is supposed to be red. 

  \subsection{Ablation study}
 
 To study the effectiveness of the different modules of our workflow, we performed different ablation experiments exploring our loss function (eq.\ref{eq:Loss_total})  and the different blocks of our network  (Fig.\ref{figMainArchi}). The results of the loss function modifications are summarized in Table \ref{Table:AblationLoss}. First, we highlighted the importance of adaptive weighting of the different $\mathcal{L}_T$ terms for the classification and segmentation tasks, with an average loss of 12.4 and 4.9 points respectively without it ($2^{nd}$ row of Table \ref{Table:AblationLoss}).  Considering this, we weighted  $\mathcal{L}_T$ taking into account the homoscedastic uncertainty of each task, following the well-known strategy of Kendall et al. \cite{kendall2018multi}. In the case the loss function is re-written as:

 \begin{align}
   \mathcal{L}_{T} &= \sum_{i=1}^3 \frac{\mathcal{L}_i}{\sigma_{i}^2} +  \sum_{i=1}^3 \log(\sigma_{i}^2)
   \label{eq:Loss_total_uncertainty}
\end{align}
 where each loss term is denoted  by $\mathcal{L}_i$ and  $\sigma_i$ is the uncertainty parameter of each task. The results show that our weighting scheme is better for each of the two scores, emphasizing the importance of learning difficult tasks after easy ones ($3^{rd}$ row of the Table \ref{Table:AblationLoss}). 
 \par We then showed the importance of image reconstruction module (Fig.\ref{figMainArchi} - block D). The deletions of the transposed CNN associated with image reconstruction and the  $\mathcal{L}_{Rec_{im}}$  term  have a significant impact on the classification and segmentation scores with a decrease of 12.1 and 20.0 points respectively  ($4^{th}$ row of Table \ref{Table:AblationLoss}). Note the detrimental effect of this ablation on the segmentation score while we used the anomaly maps from the feature maps reconstruction to determine this score. The removal of the loss term associated with feature maps reconstruction   penalizes the results with a decrease of 18.9 and 9.0 points for each of the two scores ($5^{th}$ row of the Table \ref{Table:AblationLoss}). This effect is expected since the weights of the upstream VGG network remain fixed  (Fig.\ref{figMainArchi} - block B). 
 \par To evaluate the effect of the SSL module (Fig.\ref{figMainArchi} - block A), we removed the loss term associated with the classification, as well as the data augmentation strategy.  For this experiment, the adaptive weighting scheme had to be removed because it is mainly driven by $\mathcal{L}_{cls}$. As expected, these deletions had a strong impact on the classification results with a decrease of 14.6 points while the segmentation score remains stable with a loss of 3.4 points  ($6^{th}$ row of the Table \ref{Table:AblationLoss}). Nevertheless, if the model is trained only on classification loss, there is no improvement in classification scores\footnote{For this experiment the classifications scores were calculated in function of $\mathcal{L}_{cls}$ values.} ($7^{th}$ row of the Table \ref{Table:AblationLoss}).  However, for this last experiment, the good classification results obtained on the texture objects suggest that the Cut\&Paste strategy might be unstable in our architecture.

\begin{table}[ht]
\caption{Ablation study  on loss function. The first row shows the final scores of our model, while the other rows highlight the effects of different $\mathcal{L}_T$ modifications. In each pair, the first element refers to the image-level AD ROC-AUC score and the second to the pixel-wise ROC-AUC score. The best score per column is highlighted in bold.  }
 \label{Table:AblationLoss} 
 \scalebox{0.92}{
\begin{tabular}{l|c|c|c|}
\cline{2-4}
\multicolumn{1}{c|}{\textbf{}}                                                                                                                                                               & \textbf{Mean Text.}                                        & \textbf{Mean Obj.}                  & \textbf{Mean}                                                    \\ \hline
\multicolumn{1}{|l|}{$\mathcal{L}_{T}(t) =   \alpha_{1}(t) \mathcal{L}_{cls} + \alpha_{2}(t) \mathcal{L}_{Rec_{fm}} + \alpha_{3}(t) \mathcal{L}_{Rec_{im}}$}                                    & (89.7, \textbf{88.1})                                                & (\textbf{92.2}, 92.7) & (\textbf{91.4}, \textbf{91.2}) \\
\multicolumn{1}{|l|}{$\mathcal{L}_{T} =\mathcal{L}_{cls} +\mathcal{L}_{Rec_{fm}} +\mathcal{L}_{Rec_{im}}$}                                                                                     & (65.3, 74.2)                                               & (86.7, \textbf{93.0}) & (79.0, 86.3)                                                     \\
\multicolumn{1}{|l|}{ $\mathcal{L}_{T}(t)$ equal to eq.\ref{eq:Loss_total_uncertainty} uncertainty weighting \cite{kendall2018multi}}                                                       & (97.2, 87.9)                                       & (82.53, 84.61)                        & (88.2, 85.9)                                \\
\multicolumn{1}{|l|}{$\mathcal{L}_{T}(t) =\alpha_{1}(t) \mathcal{L}_{cls} + \alpha_{2}(t)\mathcal{L}_{Rec_{fm}}$}                                                                              & (83.1, 77.3)                                               & (65.0, 80.1)                           & (71.43, 79.1)                                                    \\
\multicolumn{1}{|l|}{$\mathcal{L}_{T}(t) =\alpha_{1}(t) \mathcal{L}_{cls} + \alpha_{2}(t)\mathcal{L}_{Rec_{im}}$}                                                                              & (74.3, 74.6)                                               & (71.6, 86.0)                           & (72.5, 82.2)                                                     \\
\multicolumn{1}{|l|}{$\mathcal{L}_{T}(t) =\mathcal{L}_{Rec_{fm}}+ \mathcal{L}_{Rec_{im}}$}                                                                                                   & (63.9, 75.0)                                                & (83.2, 94.2)                            & (76.8, 87.8)                                                      \\
\multicolumn{1}{|l|}{$\mathcal{L}_{T}(t) =\mathcal{L}_{cls} $}                                                                                                                               & (\textbf{97.4}, 59.0)                                               & (75.6, 92.7)                           & (82.9, 70.6)                                                     \\
\multicolumn{1}{|l|}{$\mathcal{L}_{T}(t) =\mathcal{L}_{Rec_{fm}} $}                                                                                                                          & (62.1, 82.8)                                               & (88.3, 94.1)                           & (79.6, 90.3)                                                     
         \\ \hline
\end{tabular}
}
\end{table}

 \par To evaluate our network architecture, we first compared the performance of HaloNet as an AE, retaining only block C in  Fig.\ref{figMainArchi}, to convolutional AEs trained via $\mathcal{L}_2$ or $\mathcal{SSIM}$ loss. In this experiment, our model is optimized to reconstruct images via the combination of $\mathcal{L}_2$ and $\mathcal{SSIM}$ losses. HaloNet as an AE does not perform as well as convolutional AEs, suggesting that our model  is able to  reconstruct abnormal regions through greater generalization ability (4$^{th}$ row of Table \ref{Table:AblationArchi}). This justifies the need for the feature extractor module (Fig.\ref{figMainArchi} - block B). Next, we replaced the HaloNet AE module with the convolutional AE from DFR \cite{shi2021DFR}. In our architecture, the use of local block self-attention improves the results with an increase of 6.4 and 0.8 points for classification and segmentation respectively  (5$^{th}$ row of Table \ref{Table:AblationArchi}). 
 
 \begin{table}[ht]
 
 \centering
 \caption{Ablation study  on the architecture. The first row shows the final scores of our model. In each pair, the first element refers to the image-level AD ROC-AUC score (in percent) and the second to the pixel-wise ROC-AUC score (in percent). The best score per column is highlighted in bold.  }
 \label{Table:AblationArchi}
\begin{tabular}{l|l|l|l|}
\cline{2-4}
                                               & \multicolumn{1}{c|}{\textbf{Mean Text.}}                 & \multicolumn{1}{c|}{\textbf{Mean Obj.}}                  & \multicolumn{1}{c|}{\textbf{Mean}}                       \\ \hline
\multicolumn{1}{|l|}{HaloAE (final)}           & (\textbf{89.7}, 88.1) & (\textbf{92.2}, \textbf{92.7}) & (\textbf{91.4}, \textbf{91.2}) \\
\multicolumn{1}{|l|}{AE-l2}                    & (70.0, 69.2)                                             & (88.0, 88.9)                                             & (82.0, 82.5)                                             \\
\multicolumn{1}{|l|}{AE-SSIM}                  & (78.0, 78.2)                                             & (91.0, 91.2)                                             & (87, 86.9)                                               \\
\multicolumn{1}{|l|}{HaloAE - Block C only}    & (75.6, 67.4)                                             & (78.2, 78.8)                                             & (77.3, 75.0)                                             \\
\multicolumn{1}{|l|}{HaloaAE - Block C as CNN} & (89.5, \textbf{94.1})                                             & (82.7, 90.4)                                             & (85.0, 90.4)                                             \\ \hline
\end{tabular}
\end{table}
 
\section{Discussion and Conclusion}
To the best of our knowledge, HaloAE is the first model to incorporate a local version of Transformer, along with HaloNet \cite{vaswani2021halonet}, to handle an AD problem. Computing intra-patch correlations via the local block self-attention operation improves both detection and localization. The module optimizing the oversampling of feature maps allows us to obtain an all-in-one model, which does not require an expansive patch-based process for anomaly segmentation. We also show that the integration of an SSL approach leads to a better regularization of the AE, ultimately improving the detection score at the image level.  Finally, the improved scores brought by our new adaptive loss function weighting schemes suggest that learning multiple tasks simultaneously would be facilitated by giving increasing importance to the most difficult tasks. 
\par Overall, HaloAE performed competitively on the MVTec dataset \cite{bergmannAE2019mvtec}, with an average score of 91.4\% for image-level detection and 91.2\% for pixel-wise segmentation. The performances of our hybrid model between CNN and local Transformer suggests the importance of integrating global and local information at each step of the process. This study therefore implies that Transformer-based vision models could be improved by simultaneously applying the self-attention operation between and within patches  \cite{patel2022aggregating} \cite{liu2021swinT} \cite{han2021transformer}.
\section*{Acknowledgments}
This work for Liming Chen  was in part supported by the 4D Vision project funded by the Partner University Fund (PUF), a FACE program, as well as  the French Research Agency, l’Agence Nationale de Recherche (ANR), through the projects Learn Real ( ANR-18-CHR3-0002-01 ), Chiron (ANR-20-IADJ-0001-01),    Aristotle (ANR-21-FAI1-0009-01), and the joint support of the French national program of investment of the futur and the the regions through the PSPC FAIR Waste project. This work was granted access to the HPC resources of IDRIS under the allocation 2022-[AD011012172R1] made by GENCI.

%
%
\bibliographystyle{splncs04}
\bibliography{main}

\begin{thebibliography}{10}
\providecommand{\url}[1]{\texttt{#1}}
\providecommand{\urlprefix}{URL }
\providecommand{\doi}[1]{https://doi.org/#1}

\bibitem{adam2008robust}
Adam, A., Rivlin, E., Shimshoni, I., Reinitz, D.: Robust real-time unusual
  event detection using multiple fixed-location monitors. IEEE transactions on
  pattern analysis and machine intelligence  \textbf{30}(3),  555--560 (2008)

\bibitem{akcay2018ganomaly}
Akcay, S., Atapour-Abarghouei, A., Breckon, T.P.: Ganomaly: Semi-supervised
  anomaly detection via adversarial training. In: Asian conference on computer
  vision. pp. 622--637. Springer (2018)

\bibitem{akccay2019skipGanAnomaly}
Ak{\c{c}}ay, S., Atapour-Abarghouei, A., Breckon, T.P.: Skip-ganomaly: Skip
  connected and adversarially trained encoder-decoder anomaly detection. In:
  2019 International Joint Conference on Neural Networks (IJCNN). pp.~1--8.
  IEEE (2019)

\bibitem{baur2018VAEGAN}
Baur, C., Wiestler, B., Albarqouni, S., Navab, N.: Deep autoencoding models for
  unsupervised anomaly segmentation in brain mr images. In: International
  MICCAI Brainlesion Workshop. pp. 161--169. Springer (2018)

\bibitem{bergman2020GeomSSL}
Bergman, L., Hoshen, Y.: Classification-based anomaly detection for general
  data. arXiv preprint arXiv:2005.02359  (2020)

\bibitem{bergmannAE2019mvtec}
Bergmann, P., Fauser, M., Sattlegger, D., Steger, C.: Mvtec ad--a comprehensive
  real-world dataset for unsupervised anomaly detection. In: Proceedings of the
  IEEE/CVF Conference on Computer Vision and Pattern Recognition. pp.
  9592--9600 (2019)

\bibitem{bergmann2020ustudents}
Bergmann, P., Fauser, M., Sattlegger, D., Steger, C.: Uninformed students:
  Student-teacher anomaly detection with discriminative latent embeddings. In:
  Proceedings of the IEEE/CVF Conference on Computer Vision and Pattern
  Recognition. pp. 4183--4192 (2020)

\bibitem{bergmannSSIM2018improving}
Bergmann, P., L{\"o}we, S., Fauser, M., Sattlegger, D., Steger, C.: Improving
  unsupervised defect segmentation by applying structural similarity to
  autoencoders. arXiv preprint arXiv:1807.02011  (2018)

\bibitem{carion2020Detr}
Carion, N., Massa, F., Synnaeve, G., Usunier, N., Kirillov, A., Zagoruyko, S.:
  End-to-end object detection with transformers. In: European Conference on
  Computer Vision. pp. 213--229. Springer (2020)

\bibitem{chen2021crossvit}
Chen, C.F., Fan, Q., Panda, R.: Crossvit: Cross-attention multi-scale vision
  transformer for image classification. arXiv preprint arXiv:2103.14899  (2021)

\bibitem{chen2020iGPT}
Chen, M., Radford, A., Child, R., Wu, J., Jun, H., Luan, D., Sutskever, I.:
  Generative pretraining from pixels. In: International Conference on Machine
  Learning. pp. 1691--1703. PMLR (2020)

\bibitem{cohen2020PSVDD}
Cohen, N., Hoshen, Y.: Sub-image anomaly detection with deep pyramid
  correspondences. arXiv preprint arXiv:2005.02357  (2020)

\bibitem{cohen2020SPADE}
Cohen, N., Hoshen, Y.: Sub-image anomaly detection with deep pyramid
  correspondences. arXiv preprint arXiv:2005.02357  (2020)

\bibitem{defard2021padim}
Defard, T., Setkov, A., Loesch, A., Audigier, R.: Padim: a patch distribution
  modeling framework for anomaly detection and localization. In: International
  Conference on Pattern Recognition. pp. 475--489. Springer (2021)

\bibitem{deng2009imagenet}
Deng, J., Dong, W., Socher, R., Li, L.J., Li, K., Fei-Fei, L.: Imagenet: A
  large-scale hierarchical image database. In: 2009 IEEE conference on computer
  vision and pattern recognition. pp. 248--255. Ieee (2009)

\bibitem{devlin2018bert}
Devlin, J., Chang, M.W., Lee, K., Toutanova, K.: Bert: Pre-training of deep
  bidirectional transformers for language understanding. arXiv preprint
  arXiv:1810.04805  (2018)

\bibitem{dinh2016density}
Dinh, L., Sohl-Dickstein, J., Bengio, S.: Density estimation using real nvp.
  arXiv preprint arXiv:1605.08803  (2016)

\bibitem{dosovitskiy2020ViT}
Dosovitskiy, A., Beyer, L., Kolesnikov, A., Weissenborn, D., Zhai, X.,
  Unterthiner, T., Dehghani, M., Minderer, M., Heigold, G., Gelly, S., et~al.:
  An image is worth 16x16 words: Transformers for image recognition at scale.
  arXiv preprint arXiv:2010.11929  (2020)

\bibitem{dosovitskiy2019yoto}
Dosovitskiy, A., Djolonga, J.: You only train once: Loss-conditional training
  of deep networks. In: International conference on learning representations
  (2019)

\bibitem{golan2018GeomSSL}
Golan, I., El-Yaniv, R.: Deep anomaly detection using geometric
  transformations. arXiv preprint arXiv:1805.10917  (2018)

\bibitem{groenendijk2021alphamulti}
Groenendijk, R., Karaoglu, S., Gevers, T., Mensink, T.: Multi-loss weighting
  with coefficient of variations. In: Proceedings of the IEEE/CVF Winter
  Conference on Applications of Computer Vision. pp. 1469--1478 (2021)

\bibitem{gudovskiy2022cflow}
Gudovskiy, D., Ishizaka, S., Kozuka, K.: Cflow-ad: Real-time unsupervised
  anomaly detection with localization via conditional normalizing flows. In:
  Proceedings of the IEEE/CVF Winter Conference on Applications of Computer
  Vision. pp. 98--107 (2022)

\bibitem{han2021transformer}
Han, K., Xiao, A., Wu, E., Guo, J., Xu, C., Wang, Y.: Transformer in
  transformer. Advances in Neural Information Processing Systems  \textbf{34}
  (2021)

\bibitem{he2016resnet}
He, K., Zhang, X., Ren, S., Sun, J.: Deep residual learning for image
  recognition. In: Proceedings of the IEEE conference on computer vision and
  pattern recognition. pp. 770--778 (2016)

\bibitem{hendrycks2019geomSSL}
Hendrycks, D., Mazeika, M., Kadavath, S., Song, D.: Using self-supervised
  learning can improve model robustness and uncertainty. arXiv preprint
  arXiv:1906.12340  (2019)

\bibitem{huang2019crisscross}
Huang, Z., Wang, X., Huang, L., Huang, C., Wei, Y., Liu, W.: Ccnet: Criss-cross
  attention for semantic segmentation. In: Proceedings of the IEEE/CVF
  International Conference on Computer Vision. pp. 603--612 (2019)

\bibitem{jiang2021transgan}
Jiang, Y., Chang, S., Wang, Z.: Transgan: Two transformers can make one strong
  gan. arXiv preprint arXiv:2102.07074  \textbf{1}(3) (2021)

\bibitem{kendall2018multi}
Kendall, A., Gal, Y., Cipolla, R.: Multi-task learning using uncertainty to
  weigh losses for scene geometry and semantics. In: Proceedings of the IEEE
  conference on computer vision and pattern recognition. pp. 7482--7491 (2018)

\bibitem{khan2021Surveytransformers}
Khan, S., Naseer, M., Hayat, M., Zamir, S.W., Khan, F.S., Shah, M.:
  Transformers in vision: A survey. arXiv preprint arXiv:2101.01169  (2021)

\bibitem{li2017self}
Li, C., Yan, J., Wei, F., Dong, W., Liu, Q., Zha, H.: Self-paced multi-task
  learning. In: Thirty-First AAAI Conference on Artificial Intelligence (2017)

\bibitem{li2021cutpaste}
Li, C.L., Sohn, K., Yoon, J., Pfister, T.: Cutpaste: Self-supervised learning
  for anomaly detection and localization. In: Proceedings of the IEEE/CVF
  Conference on Computer Vision and Pattern Recognition. pp. 9664--9674 (2021)

\bibitem{liu2018classifier}
Liu, Y., Li, C.L., P{\'o}czos, B.: Classifier two sample test for video anomaly
  detections. In: BMVC. p.~71 (2018)

\bibitem{liu2021swinT}
Liu, Z., Lin, Y., Cao, Y., Hu, H., Wei, Y., Zhang, Z., Lin, S., Guo, B.: Swin
  transformer: Hierarchical vision transformer using shifted windows. arXiv
  preprint arXiv:2103.14030  (2021)

\bibitem{liznerski2020fcdd}
Liznerski, P., Ruff, L., Vandermeulen, R.A., Franks, B.J., Kloft, M.,
  M{\"u}ller, K.R.: Explainable deep one-class classification. arXiv preprint
  arXiv:2007.01760  (2020)

\bibitem{mishra2021vtadl}
Mishra, P., Verk, R., Fornasier, D., Piciarelli, C., Foresti, G.L.: Vt-adl: A
  vision transformer network for image anomaly detection and localization.
  arXiv preprint arXiv:2104.10036  (2021)

\bibitem{parmar2018imageTr}
Parmar, N., Vaswani, A., Uszkoreit, J., Kaiser, L., Shazeer, N., Ku, A., Tran,
  D.: Image transformer. In: International Conference on Machine Learning. pp.
  4055--4064. PMLR (2018)

\bibitem{patel2022aggregating}
Patel, K., Bur, A.M., Li, F., Wang, G.: Aggregating global features into local
  vision transformer. arXiv preprint arXiv:2201.12903  (2022)

\bibitem{pirnay2021inpaintTr}
Pirnay, J., Chai, K.: Inpainting transformer for anomaly detection. arXiv
  preprint arXiv:2104.13897  (2021)

\bibitem{ramachandran2019sasa}
Ramachandran, P., Parmar, N., Vaswani, A., Bello, I., Levskaya, A., Shlens, J.:
  Stand-alone self-attention in vision models. arXiv preprint arXiv:1906.05909
  (2019)

\bibitem{roth2021patchcore}
Roth, K., Pemula, L., Zepeda, J., Sch{\"o}lkopf, B., Brox, T., Gehler, P.:
  Towards total recall in industrial anomaly detection. arXiv preprint
  arXiv:2106.08265  (2021)

\bibitem{rudolphdiffernet}
Rudolph, M., Wandt, B., Rosenhahn, B.: Same same but differnet: Semi-supervised
  defect detection with normalizing flows. In: Proceedings of the IEEE/CVF
  Winter Conference on Applications of Computer Vision. pp. 1907--1916 (2021)

\bibitem{schlegl2019fAnoGan}
Schlegl, T., Seeb{\"o}ck, P., Waldstein, S.M., Langs, G., Schmidt-Erfurth, U.:
  f-anogan: Fast unsupervised anomaly detection with generative adversarial
  networks. Medical image analysis  \textbf{54},  30--44 (2019)

\bibitem{schlegl2017unsupervised}
Schlegl, T., Seeb{\"o}ck, P., Waldstein, S.M., Schmidt-Erfurth, U., Langs, G.:
  Unsupervised anomaly detection with generative adversarial networks to guide
  marker discovery. In: International conference on information processing in
  medical imaging. pp. 146--157. Springer (2017)

\bibitem{schlegl2017AnoGan}
Schlegl, T., Seeb{\"o}ck, P., Waldstein, S.M., Schmidt-Erfurth, U., Langs, G.:
  Unsupervised anomaly detection with generative adversarial networks to guide
  marker discovery. In: International conference on information processing in
  medical imaging. pp. 146--157. Springer (2017)

\bibitem{seebock2016identifying}
Seeb{\"o}ck, P., Waldstein, S., Klimscha, S., Gerendas, B.S., Donner, R.,
  Schlegl, T., Schmidt-Erfurth, U., Langs, G.: Identifying and categorizing
  anomalies in retinal imaging data. arXiv preprint arXiv:1612.00686  (2016)

\bibitem{sheynin2021localglobalTr}
Sheynin, S., Benaim, S., Polyak, A., Wolf, L.: Local-global shifting vision
  transformers  (2021)

\bibitem{shi2021DFR}
Shi, Y., Yang, J., Qi, Z.: Unsupervised anomaly segmentation via deep feature
  reconstruction. Neurocomputing  \textbf{424},  9--22 (2021)

\bibitem{simonyan2014vgg19}
Simonyan, K., Zisserman, A.: Very deep convolutional networks for large-scale
  image recognition. arXiv preprint arXiv:1409.1556  (2014)

\bibitem{sultani2018real}
Sultani, W., Chen, C., Shah, M.: Real-world anomaly detection in surveillance
  videos. In: Proceedings of the IEEE conference on computer vision and pattern
  recognition. pp. 6479--6488 (2018)

\bibitem{touvron2021Deit}
Touvron, H., Cord, M., Douze, M., Massa, F., Sablayrolles, A., J{\'e}gou, H.:
  Training data-efficient image transformers \& distillation through attention.
  In: International Conference on Machine Learning. pp. 10347--10357. PMLR
  (2021)

\bibitem{vasilev2020VAE}
Vasilev, A., Golkov, V., Meissner, M., Lipp, I., Sgarlata, E., Tomassini, V.,
  Jones, D.K., Cremers, D.: q-space novelty detection with variational
  autoencoders. In: Computational Diffusion MRI, pp. 113--124. Springer (2020)

\bibitem{vaswani2021halonet}
Vaswani, A., Ramachandran, P., Srinivas, A., Parmar, N., Hechtman, B., Shlens,
  J.: Scaling local self-attention for parameter efficient visual backbones.
  In: Proceedings of the IEEE/CVF Conference on Computer Vision and Pattern
  Recognition. pp. 12894--12904 (2021)

\bibitem{vaswani2017attention}
Vaswani, A., Shazeer, N., Parmar, N., Uszkoreit, J., Jones, L., Gomez, A.N.,
  Kaiser, {\L}., Polosukhin, I.: Attention is all you need. In: Advances in
  neural information processing systems. pp. 5998--6008 (2017)

\bibitem{venkataramanan2020attentionGAN}
Venkataramanan, S., Peng, K.C., Singh, R.V., Mahalanobis, A.: Attention guided
  anomaly localization in images. In: European Conference on Computer Vision.
  pp. 485--503. Springer (2020)

\bibitem{yang2021SAAE}
Yang, Y.: Self-attention autoencoder for anomaly segmentation  (2021)

\bibitem{zavrtanik2021reconstructionInpainting}
Zavrtanik, V., Kristan, M., Sko~caj, D.: Reconstruction by inpainting for
  visual anomaly detection. Pattern Recognition  \textbf{112},  107706 (2021)

\bibitem{zavrtanik2021draem}
Zavrtanik, V., Kristan, M., Sko{\v{c}}aj, D.: Draem-a discriminatively trained
  reconstruction embedding for surface anomaly detection. In: Proceedings of
  the IEEE/CVF International Conference on Computer Vision. pp. 8330--8339
  (2021)

\bibitem{zhang2020multi}
Zhang, Y., Gong, Y., Zhu, H., Bai, X., Tang, W.: Multi-head enhanced
  self-attention network for novelty detection. Pattern Recognition
  \textbf{107},  107486 (2020)

\bibitem{zhu2020ddter}
Zhu, X., Su, W., Lu, L., Li, B., Wang, X., Dai, J.: Deformable detr: Deformable
  transformers for end-to-end object detection. arXiv preprint arXiv:2010.04159
   (2020)

\end{thebibliography}

\newpage
\makeatletter 
\renewcommand{\thefigure}{S\arabic{figure}}
\renewcommand{\thetable}{S\arabic{table}}
\setcounter{figure}{0}
\setcounter{table}{0}

\section*{Supplementary material}
\begin{table}[h]
 \caption{Exploring the outputs of HaloAE and the post-processing procedure. The first pair score corresponds to the image-level AD ROC-AUC score in percent, and the second to the pixel-level ROC-AUC score in percent. The best score for each object is highlighted in bold. $A$ represents the anomaly map, noted $A_{im}$ or  $A_{fm}$ if calculated on the  images or on the features map, respectively (see eq.6 in main text). $A_{\{im, fm\}_{N}}$ represents the normalized anomaly map. Finnaly,  $\mathcal{N}_{filter}$ refers to the Gaussian filter applied to the normalized anomaly map.}

\begin{tabular}{l|c|c|c|c|c|}
\cline{2-6}
                                          & \textbf{$\mathcal{L}_{cls}$}      & \textbf{$A_{im_{N}}$+ $\mathcal{N}_{filter}$} & \textbf{$A_{fm}$}                  & \textbf{$A_{fm_{N}}$}               & \textbf{$A_{fm_{N}}$+ $\mathcal{N}_{filter}$}             \\ \hline
\multicolumn{1}{|l|}{\textbf{Carpet}}     & (56.9, -)                         & (\textbf{74.4}, 60.7)                            & (54.3, 88.1)                       & (60.7, \textbf{88.5})                  & (69.7, 89.4)                                              \\
\multicolumn{1}{|l|}{\textbf{Grid}}       & (\textbf{100.0}, -)  & (82.1, 53.4)                                  & (94.5, 82.7)                       & (95.2, 83.0)                        & (95.1, \textbf{83.1})                        \\
\multicolumn{1}{|l|}{\textbf{Leather}}    & (71.0, -)                         & (60.2, 78.3)                                  & (97.2, 98.0)                       & (\textbf{97.8}, 98.1)  & (\textbf{97.8}, \textbf{98.5})  \\
\multicolumn{1}{|l|}{\textbf{Tile}}       & (51.5, -)                         & (92.6, 66.1)                                  & (93.3, 75.9)                       & (95.2, 76.1)                        & (\textbf{95.7}, \textbf{78.5} ) \\
\multicolumn{1}{|l|}{\textbf{Wood}}       & (93.2, -)                         & (99.0, 77.4)                                  & (99.7, 90.7)                       & (99.9, 90.3)                        & (\textbf{100.0}, \textbf{91.1}) \\ \hline
\multicolumn{1}{|l|}{\textbf{Mean Text.}} & (74.5, -)                         & (81.66, 67.2)                                 & (87.8, 87.1)                       & (\textbf{89.8}, 87.2)  & (89.7, \textbf{88.1})                        \\ \hline
\multicolumn{1}{|l|}{\textbf{Bottle}}     & (98.4, -)                         & (99.9, 86.7)                                  & (99.9, 90.0)                       & (\textbf{100.0}, 91.7) & (\textbf{100.0}, \textbf{91.9}) \\
\multicolumn{1}{|l|}{\textbf{Cable}}      & (\textbf{100.0}, -)  & (62.8, 76.3)                                  & (79.2, 77/9)                       & (84.6, 86.1)                        & (84.6, \textbf{87.6})                        \\
\multicolumn{1}{|l|}{\textbf{Capsule}}    & (\textbf{96.8}, -)   & (54.5, 63.6)                                  & (83.2, 97.3)                       & (88.4, 97.4)                        & (88.4, \textbf{97.8})                        \\
\multicolumn{1}{|l|}{\textbf{HazelNut}}   & (99.4, -)                         & (86.3, 76.0)                                  & (98.9, 97.9)                       & (99.6, 97.7)                        & (\textbf{99.8}, \textbf{97.8})  \\
\multicolumn{1}{|l|}{\textbf{MeatalNut}}  & (\textbf{98.0}, -)   & (65.2, 69.2)                                  & (85.6, 86.3)                       & (88.4, 84.5)                        & (88.4, \textbf{85.2})                        \\
\multicolumn{1}{|l|}{\textbf{Pill}}       & (\textbf{100.0},  -) & (50.8, 77.2)                                  & (86.4, \textbf{92.8}) & (90.6, 89.9)                        & (90.1, 91.5)                                              \\
\multicolumn{1}{|l|}{\textbf{Screw}}      & (\textbf{100.0}, -)  & (54.6, 78.5)                                  & (88.6, 98.8)                       & (89.6, 98.6)                        & (89.6, \textbf{99.0})                        \\
\multicolumn{1}{|l|}{\textbf{Toothbrush}} & (58.1. -)                         & (89.7, 81.0)                                  & (94.7, 93.0)                       & (\textbf{97.2}, 92.6)  & (\textbf{97.2}, \textbf{92.9})  \\
\multicolumn{1}{|l|}{\textbf{Transistor}} & (\textbf{92.3}, -)   & (81.5, 79.9)                                  & (80.0, 84.8)                       & (84.4, 85.6)                        & (84.4, 87.5)                                              \\
\multicolumn{1}{|l|}{\textbf{Zipper}}     & (51.4, -)                         & (\textbf{99.7}, 86.8)            & (99.7, 95.4)                       & (\textbf{99.7}, 95.3)  & (\textbf{99.7}, \textbf{96.0})  \\ \hline
\multicolumn{1}{|l|}{\textbf{Mean Obj.}}  & (89.4, -)                         & (74.5, 77.5)                                  & (89.6, 91.6)                       & (\textbf{92.3}, 91.9)  & (92.2, \textbf{92.7})                        \\ \hline
\multicolumn{1}{|l|}{\textbf{Mean}}       & (84.4, -)                         & (76.9, 74.1)                                  & (89.0, 90.0)                       & (\textbf{91.4}, 90.4)  & (\textbf{91.4},  \textbf{91.2}) \\ \hline
\end{tabular}
\label{supTableDetailScores}
\end{table}

 \begin{figure}[h]
 \centering
 \includegraphics[width=.8\textwidth]{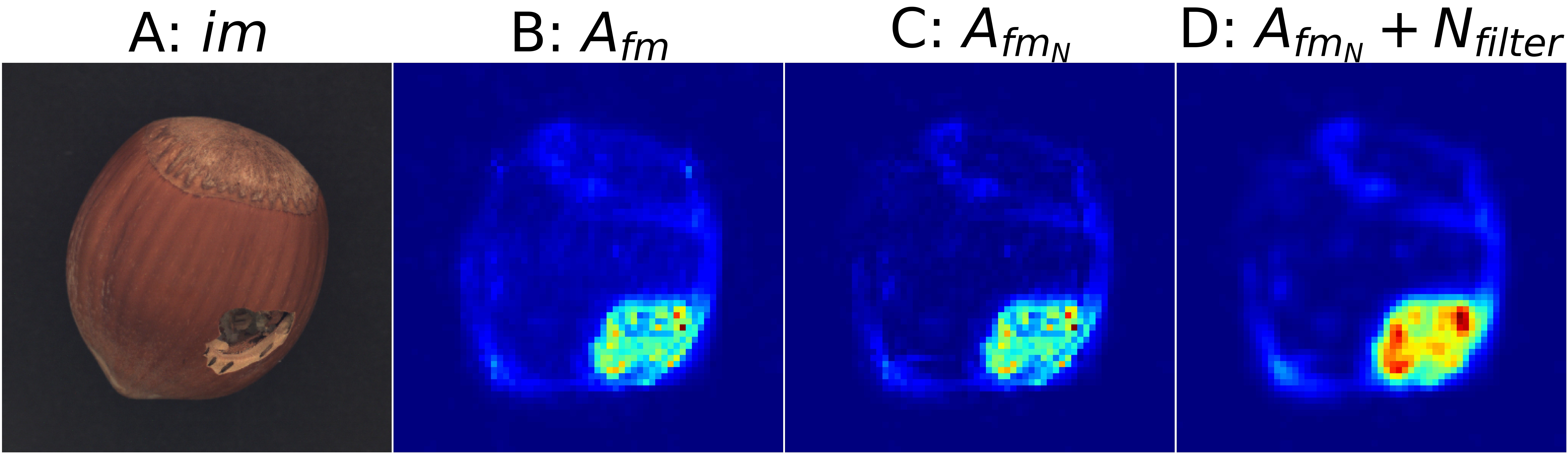}
 \caption{Post processing workflow. \textbf{A)} Input image. \textbf{B)} Anomaly map (see eq.6 in main text). \textbf{C)} Normalized anomaly map (see eq.7 in main text). \textbf{D)} Normalized anomaly map smoothed with a Gaussian filter.}  
 \label{figPostProcess}
\end{figure}

 \begin{figure}[h]
 \centering
 \includegraphics[width=
 \textwidth]{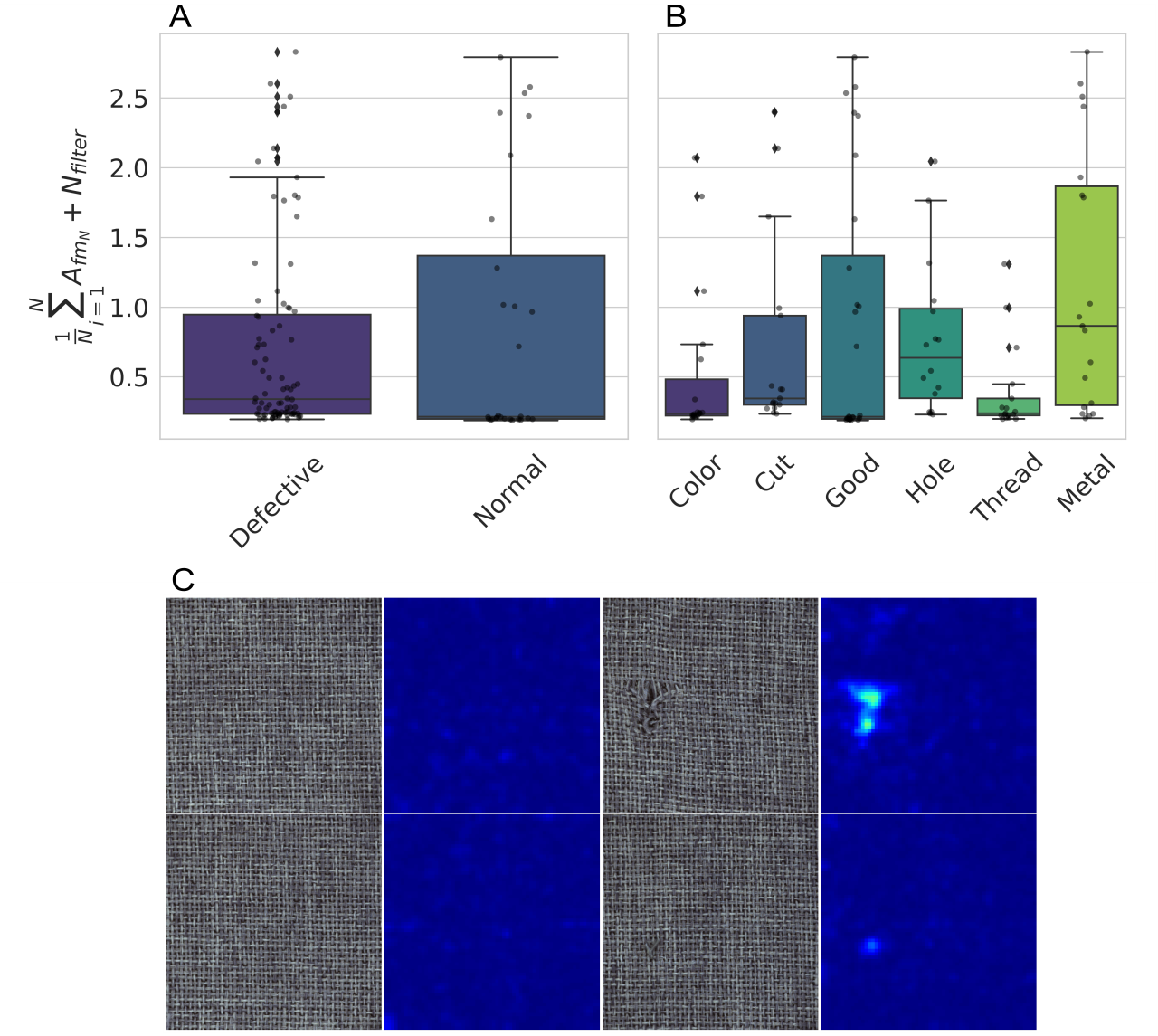}
 \caption{Classification results on carpet. \textbf{A)} Distribution of means of post-processed anomaly maps computed on the feature map, for defect free and abnormal objects. \textbf{B)} Distribution of means of post-processed anomaly maps computed on the feature map by defect category. Defect-free objects and anomalous objects have similar distributions.  \textbf{C)} Carpet anomaly map, on the left objects without defects, on the right abnormal objects.  }  
 \label{figResCarpet}
\end{figure}

 \begin{figure}[h]
 \centering
 \includegraphics[width=
 \textwidth]{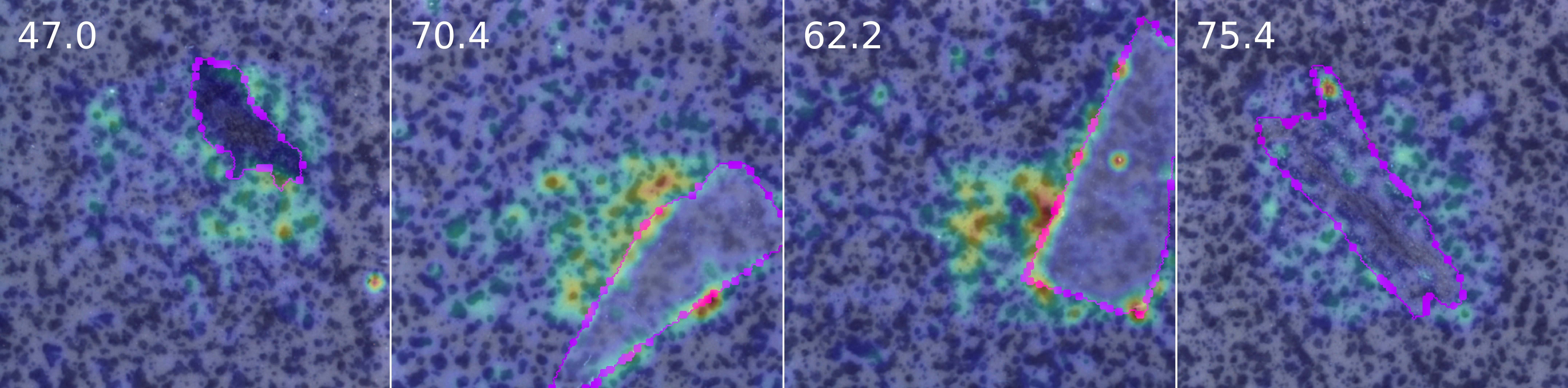}
 \caption{Segmentation results on tiles. Each image shows the segmentation anomaly maps computed on the feature maps, the ground truth location is surrounded by a pink line.  For each image, the anomaly score per pixel in percentage is shown in the upper left corner. }  
 \label{figResTile}
\end{figure}

\end{document}